\documentclass{article}
\usepackage{spconf,amsmath,graphicx}

\usepackage{bbm}
\usepackage{xcolor}
\usepackage{caption}
\usepackage{subcaption}
\usepackage{url}
\usepackage{amssymb}
\usepackage{bm}
\usepackage{amsfonts}
\graphicspath{ {.} }
\newcommand{\gtlt}{{\left. \begin{array}{ll} H_0  \\ >   \\ <   \\ H_1  \\ \end{array} \right. }}

\title{ADVERSARIALLY ROBUST CLASSIFICATION BASED ON GLRT}
\name{Bhagyashree Puranik, Upamanyu Madhow, Ramtin Pedarsani\thanks{This work was supported by the Army Research Office, and the National Science Foundation grant CCF 1909320.}}
\address{University of California Santa Barbara}
\begin{document}
%
\maketitle
\begin{abstract}

Machine learning models are vulnerable to adversarial attacks that can often cause misclassification by introducing small but well designed perturbations. In this paper, we explore, in the setting of classical composite hypothesis testing, a defense strategy based on the generalized likelihood ratio test (GLRT), which jointly estimates the class of interest and the adversarial perturbation. We evaluate the GLRT approach for the special case of binary hypothesis testing in white Gaussian noise under $\ell_{\infty}$ norm-bounded adversarial perturbations, a setting for which a minimax strategy optimizing for the worst-case attack is known. We show that the GLRT approach yields performance competitive with that of the minimax approach under the worst-case attack, and observe that it yields a better robustness-accuracy trade-off under weaker attacks, depending on the values of signal components relative to the attack budget. We also observe that the GLRT defense generalizes naturally to more complex models for which optimal minimax classifiers are not known.

\end{abstract}

\section{Introduction}
\label{sec:intro}

Machine learning models such as deep neural networks and regression methods have become pervasively deployed in large-scale commercial applications that are safety-critical, such as facial recognition for surveillance, autonomous driving and virtual assistants. It has been shown that an adversary is often able to add small perturbations to signals in an intelligent way to cause misclassification with high confidence~\cite{szegedy,rolli}. In applications that demand robustness in machine learning methods, adversarial attacks are fundamental threats. There have been several defense mechanisms suggested, followed by proposal of stronger adversaries to circumvent the defenses~\cite{carlini2017,carlini2018}. A state-of-the-art defense~\cite{madry_iclr2018} against such attacks is to train with adversarial examples--this is purely empirical and cannot provide robustness guarantees or insights.

In this paper, we seek fundamental insight by investigating adversarial classification in the setting of classical hypothesis testing, in which the class-conditional distributions of the data is known. We propose the well-known generalized likelihood ratio test (GLRT) as a general approach to defense, in which the desired class and the action of the adversary (viewed as a nuisance parameter) are estimated jointly.  The GLRT approach is general, since it applies to any composite hypothesis testing problem~\cite{poor}, unlike minimax strategies optimizing for worst-case attacks, which are difficult to find.  We compare the GLRT and minimax approaches for a simple setting, binary Gaussian hypothesis testing with $\ell_{\infty}$ bounded attacks, for which the minimax strategy has been recently derived \cite{bhagoji}.  We show that the proposed GLRT approach provides competitive robustness guarantees when the attacker employs the full attack budget, while providing better robustness-accuracy trade-off for weaker attacks.

\vspace{0.25cm}
\noindent
\textbf{Related Work:} There is a growing body of research on coming up with provable robustness guarantees against adversarial attacks~\cite{liang1,liang2,wong,wong2,sinha,mirman,hein,cisse,marzi}. A recent paper \cite{hassani} addresses the problem of finding optimal robust classifiers in a binary classification problem, with the class conditional distributions possessing symmetric means and white Gaussian noise. Optimal robust classifiers are derived for binary and ternary classification problems when the perturbations are $\ell_2$ norm-bounded. For the case when perturbations are $\ell_{\infty}$ norm bounded, they restrict attention to the class of linear classifiers and then obtain optimum robust linear classifiers for two and three-class classification problems. In general, finding robust optimal classifiers for $\ell_{\infty}$ norm bounded adversarial perturbations is not easily tractable. Analytical results have been shown only for special cases, such as in \cite{bhagoji}, where optimal robust classifiers and optimal adversarial risk are characterized in binary classification setting under Gaussian models with symmetric means, same covariance matrices and uniform priors.

\section{GLRT-based defense}
\label{sec:glrt_des}

Throughout the paper, we represent vectors in boldface letters and scalars in regular letters. The norm $|| \cdot ||$ denotes $\ell_2$ norm unless specified otherwise. Consider the following standard classification or hypothesis testing problem: $\mathcal{H}_k: \mathbf{X} \sim p_k(\mathbf{x})$. The presence of an adversary increases the uncertainty about the class-conditional densities, which can be modeled as a composite hypothesis testing problem:
\begin{equation*}
\mathcal{H}_k: \mathbf{X} \sim p_{\theta}(\mathbf{x}), \theta \in \Theta_k,
\end{equation*}
where the size of the uncertainty sets $\Theta_k$ depends on the constraints on the adversary. The GLRT defense consists of joint maximum likelihood estimation of the class and the adversary's parameter:
	\begin{equation*}
	\hat{k} = \arg\max\limits_{k} \max\limits_{\theta \in \Theta_k} {\it{p}}_{\theta}(\mathbf{x}).
	\end{equation*}

\noindent
{\bf Gaussian hypothesis testing:}	
We now apply this framework to Gaussian hypothesis testing with an adversary which can add an $\ell_{\infty}$-bounded perturbation $\mathbf{e}$:
 $|| \mathbf{e} ||_{\infty} \leq \epsilon$, where we term $\epsilon$ the ``attack budget'' or ``adversarial budget''.
\begin{equation*}
	\mathcal{H}_k: \mathbf{X} = \bm{\mu}_k + \mathbf{e} + \mathbf{N},
	\end{equation*}
where $\mathbf{X} \in \mathbb{R}^d$, $\mathbf{N}\sim \mathcal{N}(0, \sigma^2 I_d)$ is white Gaussian noise. We assume
that the adversary has access to the true hypothesis and knows the distributions under each of these hypotheses. 

Conditioned on the hypothesis $k$ and the perturbation $\mathbf{e}$, the negative log likelihood is a standard quadratic expression.  Applying GLRT, we first estimate $\mathbf{e}$ under each hypothesis:
\begin{equation*}
	\hat{\mathbf{e}}_k = \arg\min\limits
	_{\mathbf{e}: ||\mathbf{e}||_{\infty} \leq \epsilon} ||\mathbf{X} - \bm{\mu}_k - \mathbf{e}||^2.
\end{equation*}
and then plug in to obtain the cost function to be minimized over $k$: 
\begin{equation} \label{Ck_1}
C_k =  ||\mathbf{X} - \bm{\mu}_k - \hat{\mathbf{e}}_k||^2
\end{equation}

This yields intuitively pleasing answers in terms of the symmetric ReLU $g_{\epsilon}(x) = \text{sign}(x) \text{max} \left( 0, |x|-\epsilon \right)$ and 
its ``complement'', $f_{\epsilon}(x) = x - g_{\epsilon}(x)$. 
The estimated perturbation under hypothesis $k$ is obtained as
$\hat{\mathbf{e}}_k = f_{\epsilon}\left( \mathbf{X} - \bm{\mu}_k \right)$, where the non-linearity is applied coordinate-wise.  Substituting into (\ref{Ck_1}),
we obtain 
\begin{equation} \label{Ck_2}
C_k = ||g_{\epsilon} \left( \mathbf{X} - \bm{\mu}_k \right) ||^2
\end{equation}
where the double-sided ReLU is applied coordinate-wise.  Thus, the GLRT detector 
\begin{equation*}
\hat{k} = \arg\min_k C_k
\end{equation*}
is a modified version of the standard minimum distance rule where the coordinate-wise differences between the observation and the template are passed
through a double-sided ReLU.

\vspace{0.25cm}
\noindent
{\bf Minimax formulation:} An alternative to the GLRT defense, which treats adversarial perturbation as a ``nuisance parameter'' is a game-theoretic formulation. 
Let $\mathcal{H}$ denote the true hypothesis and $\mathcal{\hat{H}}$ be a classifier. The adversary attempts to maximize the probability of error by choosing a suitable perturbation, while the defender tries to choose a classifier such that the expected probability of error is minimized. We consider the perturbations $\mathbf{e}: ||\mathbf{e}||_{\infty} \leq \epsilon$. Thus the optimum adversarial risk is:
	\begin{equation*}
	R^* = \min\limits_{\mathcal{\hat{H}}} \mathbf{\mathbb{E}}\big[\sup\limits_{\mathbf{e} : ||\mathbf{e}||_{\infty} \leq \epsilon } \mathbbm{1}(\mathcal{\hat{H}} \neq \mathcal{H}) \big].
	\end{equation*}
Clearly, this is the best possible approach for defending against worst-case attacks.  Unfortunately, such minimax games are difficult to solve, unlike the more generally applicable GLRT approach. Furthermore, the optimal minimax solution
may be overly conservative, unnecessarily compromising performance against attacks that are weaker than, or different from, the worst-case attack.
In such scenarios, we expect the GLRT approach, which estimates the attack parameters, to provide an advantage.
In order to compare the minimax and GLRT approaches, for the remainder of this paper, 
we specialize to a setting where the minimax solution is known: binary Gaussian hypothesis testing with symmetric means and equal priors.

\section{Binary Gaussian Hypothesis Testing}
\label{sec:analysis}

We now focus on the binary hypothesis testing problem with equal priors for which the minimax rule is known \cite{bhagoji}:
	\begin{eqnarray*}
		\mathcal{H}_{0}&:& \mathbf{X} = \bm{\mu} + \mathbf{e} +\mathbf{N}, \\
		\mathcal{H}_{1}&:& \mathbf{X} = -\bm{\mu} + \mathbf{e} +\mathbf{N},
	\end{eqnarray*}
where $\mathbf{e}$ is chosen by an  $\ell_{\infty}$ bounded adversary, with adversarial budget $\epsilon$, who knows the true hypothesis. In the absence of attack, the optimal rule is a minimum distance rule, which can be alternatively written as a linear detector:
\begin{equation*} \label{linear}
\mathbf{w}^T  \mathbf{X} \gtlt 0
\end{equation*}
where $\mathbf{w}_{clean} = \bm{\mu}$ or any positive scalar multiple of it. Under uniform priors, it is shown in \cite{bhagoji} that the minimax decision rule is also a linear detector, with
$\mathbf{w}_{minimax} = g_{\epsilon} \left( \bm{\mu} \right)$. The worst-case attack is $\mathbf{e} = -\epsilon.\text{sign}(\bm{\mu})$ under $\mathcal{H}_0$ and 
$\mathbf{e} = \epsilon. \text{sign}(\bm{\mu})$ under $\mathcal{H}_1$.

Under this attack, it is easy to see that the ``defenseless'' linear detector $\mathbf{w}_{clean}$ makes errors with probability at least half whenever the attack budget satisfies
$\epsilon > ||  \bm{\mu} ||^2/||  \bm{\mu} ||_1$. Thus, the system is less vulnerable (i.e., the adversary needs a large attack budget) when the $\ell_1$ norm of $ \bm{\mu}$  is small 
relative to the $\ell_2$ norm.  That is, signal sparsity helps in robustness, as has been observed before~\cite{bakiskan, marzi}.

The minimax rule derived in \cite{bhagoji} applies a double-sided ReLU to the ``signal template'' $\bm{\mu}$. Thus, it simply ignores signal coordinates whose sign could be flipped using the worst-case attack budget, and shrinks the remaining coordinates to provide an optimal rule {\it assuming that the worst-case attack has been applied.} 
Comparing with the GLRT rule
\begin{equation} \label{glrt_binary}
C_1 = ||g_{\epsilon} \left( \mathbf{X} + \bm{\mu} \right) ||^2 \gtlt C_0 = ||g_{\epsilon} \left( \mathbf{X} - \bm{\mu} \right) ||^2
\end{equation}
we see that GLRT applies the (coordinate-wise) double-sided ReLU to the difference between the observation and signal templates, and hence should be better able to adapt to the attack level (as long as it is smaller than the budget $\epsilon$).  

\subsection{Analysis}

Since the GLRT rule is nonlinear, its performance is more difficult to characterize than that of a linear detector. However, we are able to provide insight via a central limit
theorem (CLT) based approximation (which is accurate for moderately large dimension $d$). By the symmetry of the observation model (and the resulting symmetry induced on the attack model), we may condition on $\mathcal{{H}}_0$ and the
corresponding attack $\mathbf{e} = -\epsilon.\text{sign}(\bm{\mu})$, and consider $\mathbf{X} = \bm{\mu} - \epsilon \text{sign}(\bm{\mu}) + \mathbf{N}$. The costs are
\begin{eqnarray*}
C_0 &=& \sum\limits_{i = 1}^d(g_{\epsilon}(-\epsilon \text{sign}(\bm{\mu}[i]) + \mathbf{N}[i]))^2\\ 
C_1 &=& \sum\limits_{i = 1}^d(g_{\epsilon}(2\bm{\mu}[i]-\epsilon \text{sign}(\bm{\mu}[i]) + \mathbf{N}[i]))^2,
\end{eqnarray*}
and the error probability of interest is 
\begin{equation}
P_e = P_{e|0} = P(C = C_1 - C_0 < 0|\mathcal{{H}}_0).
\end{equation}

We now perform a coordinate-wise analysis of the cost difference $C[i] = C_1[i] - C_0[i]$, denoting its mean by $m_i$ and variance by $\rho_i^2$, and then applying CLT on the sum across coordinates. The error probability is then estimated as:
\begin{equation}
P_e = P_{e|0} = P\big(\sum_{i = 1}^{d}C[i] < 0\big) \approx Q \left( \frac{\sum_{i=1}^d m_i}{\sqrt{\sum_{i=1}^d \rho_i^2}} \right)
\label{eqn:clt}
\end{equation}
The approximate equality in (\ref{eqn:clt}) can be formalized to exact equality in the limit under the mild assumption of satisfying Lindeberg's condition for CLT to hold for independent, but not necessarily identically distributed random variables.

Consider a particular coordinate $i$, set $C = C[i]$, and let
$\bm{\mu} [i] = \mu$.  Assume $\mu > 0$ without loss of generality: we simply replace $\mu$ by $| \mu |$ after performing our analysis, since 
the analysis is entirely analogous for $\mu < 0$, given the symmetry of the noise and the attack. 
We can numerically compute the mean and variance of the cost difference for the coordinate, $C = \left( g_{\epsilon}(2\mu + N - \epsilon) \right)^2  - \left( g_{\epsilon}(N - \epsilon)\right)^2$, but the following lower bound yields insight: 
\begin{equation}
C  \geq Y \triangleq \mathbbm{1}_{\{N \geq -t \}} (t + N )^2 - N^2
\end{equation}
where $t = 2( \mu - \epsilon )$. Note that $t > 0$ ($| \mu | > \epsilon$) corresponds to coordinates that
the minimax detector would retain. The high-SNR ($t/\sigma$ large) behavior is interesting.
For $t > 0$, we can show that $Y \approx t^2 + 2Nt$; these coordinates exhibit behavior similar to the minimax detector.
On the other hand, for $t < 0$, $Y \approx - N^2$; these coordinates, which would have been deleted by the minimax detector, contribute noise in favor of
the incorrect hypothesis (this becomes negligible at high SNR).  These observations can be used to show that, at high SNR, the performance of the GLRT detector approaches that of the minimax detector under worst-case attack.

\begin{figure}[t!]
	\centering
	\includegraphics[width=\columnwidth]{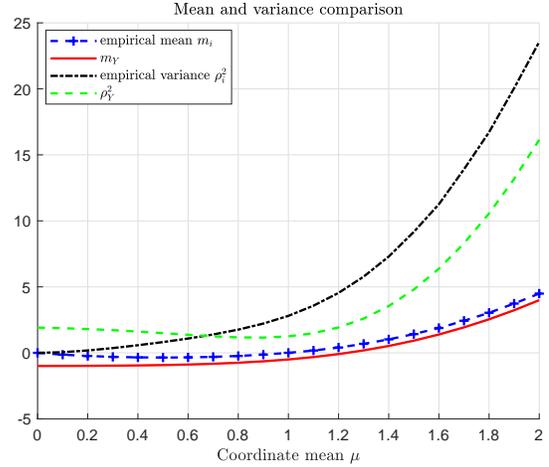}
	\caption{Comparison of empirical mean and variance of $C[i]$ with the calculated mean and variance of the lower bounding variable $Y_i$.}
	\label{fig:coord_mean_var}
\end{figure}

Without loss of generality, let us redefine $t = 2(|\mu| - \epsilon)$. The mean and variance of $Y$, irrespective of $\text{sign}(\mu)$, can be computed in closed form as follows
\begin{eqnarray}
	m_Y &=&  Q\Big(\frac{-t}{\sigma}\Big) (t^2 + \sigma^2) - \sigma^2 + \sigma t N\Big(\frac{t}{\sigma}; 0,1\Big)\label{eqn:mean}\\
	\rho^2_Y&=& 3\sigma^4 + Q\Big(\frac{-t}{\sigma}\Big) (t^4 + 4t^2\sigma^2 - 3\sigma^4)\nonumber\\
	&& + \sigma t N(t/\sigma; 0,1)(t^2 + 3\sigma^2) - m_Y^2, \label{eqn:var}
\end{eqnarray}
where $N(.;0,1)$ denotes the density of standard Gaussian (zero-mean, unit-variance) random variable, and $Q(.)$ its complementary CDF. Figure~\ref{fig:coord_mean_var} shows the empirical mean and empirical variance of $C[i]$, i.e., $m_i$ and $\rho^2_i$,  in comparison with $m_Y$ and $\rho^2_{Y}$ obtained through (\ref{eqn:mean}) and (\ref{eqn:var}). Here, the adversarial budget is set to $\epsilon = 1$ and noise variance $\sigma^2 = 1$. 

The probability of error in (\ref{eqn:clt}) can also be bounded by applying CLT on the lower bounding terms $Y_i \leq C[i]$ as follows
\begin{eqnarray*}
P\big(\sum_{i = 1}^{d}C[i] < 0\big) \leq P\big(\sum_{i = 1}^{d}Y_i < 0\big) \approx Q \left( \frac{\sum_{i=1}^d m_{Y_i}}{\sqrt{\sum_{i=1}^d \rho^2_{Y_i}}} \right).
\end{eqnarray*}

Bounding the probability of error in this fashion helps in yielding the following insight. Under low noise limit ($\sigma^2 \rightarrow 0$), the variance $\rho^2_{Y_i} = 0, \forall i$; and the mean is given by $m_{Y_i} = t^2$, if $|\bm{\mu}[i]| > \epsilon$, otherwise it is zero. Thus as long as  $\exists i$ such that $|\bm{\mu}[i]| > \epsilon$, we have $P_e = 0$. Also note that since each of the means and variances are $\mathcal{O}(1)$ terms, we have
$P_e \leq k_1 e^{-k_2 d}$, where $k_1$, $k_2$ are positive constants.

\section{Numerical Examples}
\begin{figure}[t!]
	\centering
	\includegraphics[width=\columnwidth]{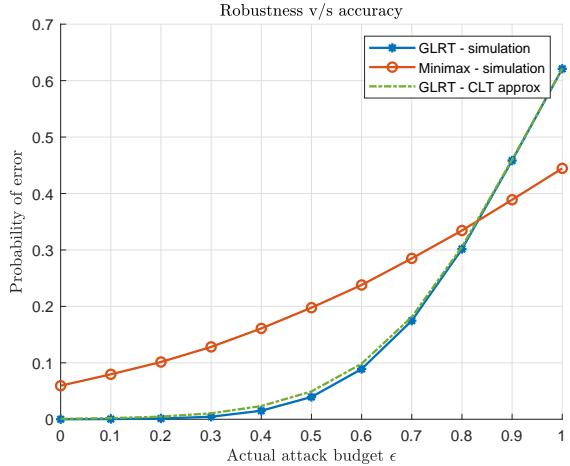}
	\caption{Error probability comparison as actual attack is varied, while the designed adversarial budget is fixed to $\epsilon_{des} =1$. Also shown is the error probability as approximated by employing CLT.}
	\label{fig:rob_acc}
\end{figure}

We consider the following example realization of the binary classification problem with uniform priors to draw a comparison with the minimax optimal scheme. We represent the designed adversarial budget as $\epsilon_{des} = 1$ and the actual attack is of the form to $e = \mp\epsilon \text{sign}(\bm{\mu})$, where $\epsilon$ is varied from $0$ to $\epsilon_{des}$. A fraction $p = 0.1$ of the $d = 20$ coordinates are such that $\mu = 1.1\epsilon_{des}$ and the rest with $\mu = 0.9 \epsilon_{des}$. Figure~\ref{fig:rob_acc} depicts the simulation results for these parameters with noise variance $\sigma^2 = 1$. For this example, the GLRT scheme has lower probability of error than the minimax scheme for weaker attacks. The error probability calculated by applying CLT to the conditional cost statistics as in (\ref{eqn:clt}), is also plotted to show that the estimates are close under reasonably high number of dimensions.

\begin{figure}[t!]
	\centering
	\includegraphics[width=\columnwidth]{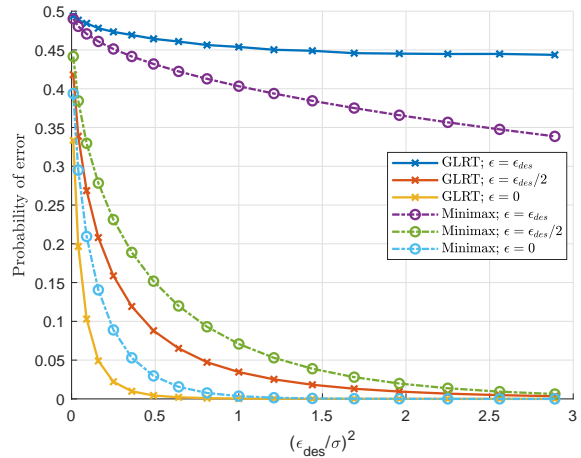}
	\caption{Predicted probability of error for the two detectors as a function of $(\epsilon_{des}/\sigma)^2$ for different values of actual attack budgets.}
	\label{fig:prob_err}
\end{figure}

Let us suppose that the dimension $d$ is large enough. Let a fraction $p$ of the coordinates have signal components $\mu = a\epsilon_{des}$ and a fraction $(1-p)$ have $\mu = b\epsilon_{des}$, where $a>1$ and $0\leq b\leq 1$. Let the designed adversarial budget be $\epsilon_{des}$ and the actual attack be $\mathbf{e} = -\epsilon \text{sign}(\bm{\mu})$, where $\epsilon=k \epsilon_{des}$, ($k \leq 1$). The effective signal-to-noise ratio (SNR) for the minimax and GLRT detectors are as follows:
\begin{equation*}
\text{SNR}_{\text{minimax}} = (a - k)^2 dp\Big(\frac{\epsilon_{des}}{\sigma}\Big)^2
\end{equation*}
\begin{equation*}
\text{SNR}_{\text{GLRT}} = d \frac{(p m_a + (1-p)m_b)^2}{p \rho^2_a + (1-p) \rho^2_b}
\end{equation*}
where $m_a$ and $m_b$ are the means, $\rho_a^2$ and $\rho_b^2$ are the variances of a single coordinate $C[i]$ contributed by terms with components $a\epsilon_{des}$ and $b\epsilon_{des}$ respectively. The probability of error in both of these cases is given by $Q(\sqrt{\text{SNR}})$ and the same is plotted in Figure~\ref{fig:prob_err} against $(\epsilon_{des}/\sigma)^2$, for different values of actual attack budget $\epsilon$, for a problem instance with parameters $d = 20$, $p = 0.3$, $\epsilon_{des} = 1$, $a = 1.1$ and $b = 0.9$. At higher noise levels, GLRT outperforms minimax scheme for weaker attacks.

\section{Conclusion}

The GLRT approach to robust hypothesis testing explored in this paper can be generalized to complex models, in contrast to the difficulty of finding optimal minimax classifiers. For the simple model considered here, for which the minimax detector is known, we show that the GLRT detector has the same asymptotic performance as the minimax detector at high SNR for $\ell_{\infty}$ bounded adversarial perturbations at a designated attack level. For attack levels lower than this designated level, the GLRT detector can provide better performance, depending on the specific values of the signal components relative to the attack budget.

An interesting direction for future research is to apply the GLRT approach to more complex data and attack models. It is also of interest to explore the minimax formulation in such settings: even if it is difficult to find the optimal minimax rule, a combination of insights from the minimax and GLRT formulations for simpler models might be useful.

\vfill\pagebreak

\bibliographystyle{IEEEbib}
\bibliography{strings,refs}

\end{document}